\documentclass[11pt,a4paper]{article}
\usepackage[hyperref]{emnlp-ijcnlp-2019}
\usepackage{times}
\usepackage{latexsym}
\usepackage{graphicx} 
\usepackage{url}
\usepackage{amsmath}
\usepackage{soul}

\aclfinalcopy 

\title{Unsupervised Question Answering for Fact-Checking}

\author{Mayank Jobanputra \\
  Department of Computer Science \& Engineering \\
  IIIT Delhi \\
  New Delhi, India \\
  {\tt mayankj@iiitd.ac.in}}
\date{}
\begin{document}
\maketitle
\begin{abstract}
 Recent Deep Learning (DL) models have succeeded in achieving human-level accuracy on various natural language tasks such as question-answering, natural language inference (NLI), and textual entailment. These tasks not only require the contextual knowledge but also the reasoning abilities to be solved efficiently. In this paper, we propose an unsupervised question-answering based approach for a similar task, fact-checking. We transform the FEVER dataset into a Cloze-task by masking named entities provided in the claims. To predict the answer token, we utilize pre-trained Bidirectional Encoder Representations from Transformers (BERT). The classifier computes label based on the correctly answered questions and a threshold. Currently, the classifier is able to classify the claims as ``SUPPORTS'' and ``MANUAL\_REVIEW''. This approach achieves a label accuracy of 80.2\% on the development set and 80.25\% on the test set of the transformed dataset.
 
\end{abstract}

\section{Introduction}

Every day textual information is being added/updated on Wikipedia, as well as other social media platforms like Facebook, Twitter, etc. These platforms receive a huge amount of unverified textual data from all its users such as News Channels, Bloggers, Journalists, Field-Experts which ought to be verified before other users start consuming it. This information boom has increased the demand of information verification also known as Fact Checking. Apart from the encyclopedia and other platforms, domains like scientific publications and e-commerce also require information verification for reliability purposes. Generally, Wikipedia authors, bloggers, journalists and scientists provide references to support their claims. Providing referenced text against the claims makes the fact checking task a little easier as the verification system no longer needs to search for the relevant documents.

Wikipedia manages to verify all this new information with a number of human reviewers. Manual review processes introduce delays in publishing and is not a well scalable approach. To address this issue, researchers have launched relevant challenges, such as the Fake News Challenge  (\citealp{pomerleau2017fake}), Fact Extraction and VERification (FEVER) (\citealp{thorne2018fever}) challenge along with the datasets. Moreover, Thorne and Vlachos (\citeyear{thorne2018automated}) released a survey on the current models for automated fact-checking. FEVER is the largest dataset and contains around 185k claims from the corpus of 5.4M Wikipedia articles. The claims are labeled as ``SUPPORTS'', ``REFUTES'', or ``NOT ENOUGH INFO'', based on the evidence set. 

In this paper, we propose an unsupervised question-answering based approach for solving the fact-checking problem. This approach is inspired from the memory-based reading comprehension task that humans perform at an early age. As we know that kids in schools, first read and learn the syllabus content so that they can answer the questions in the exam. Similarly, our model learns a language model and linguistics features in unsupervised fashion from the provided Wikipedia pages.

To transform the FEVER dataset into the above-mentioned task, we first generate the questions from the claims. In literature, there are majorly two types of Question Generation systems: Rule-based and Neural Question Generation (NQG) model based. Ali et al. (\citeyear{ali2010automation}) proposed a rule-based pipeline to automate the question generation using POS (Part-of-speech) tagging and Named Entity Recognition (NER) tagging from the sentences. Recently, many NQG models have been introduced to generate questions in natural language. Serban et al. (\citeyear{serban2016generating}) achieved better performance for question generation  utilizing (passage, question, answer) triplets as training data and an encoder-decoder based architecture as their learning model.

Du et al. (\citeyear{du2017learning}) introduced a sequence-to-sequence model with an attention mechanism, outperforming rule-base question generation systems. Although the models proposed in (\citealp{kim2019improving}; \citealp{wang2017bilateral}) are effective, they require a passage to generate the plausible questions which is not readily available in the FEVER dataset. To resolve the issues and to keep the system simple but effective, we chose to generate questions similar to a Cloze-task or masked language modeling task. Such a task makes the problem more tractable as the masked entities are already known (i.e. named entities) and tight as there is only one correct answer for a given question. Later when the answers are generated, due to the question generation process, it becomes very easy to identify the correct answers.
 
We use the BERT's (Bidirectional Encoder Representations from Transformers) (\citealp{devlin2018bert}) masked language model, that is pre-trained on Wikipedia articles for predicting the masked entities. Currently, neither the claim verification process nor the question generation process mandates explicit reasoning. For the same reason, it is difficult to put ``REFUTES'' or ``NOT ENOUGH INFO'' labels. To resolve this issue, we classify the unsupported claims as ``MANUAL\_REVIEW'' instead of labeling them as ``NOT ENOUGH INFO'' or ``REFUTES''. 

In the literature, the shared task has been tackled using pipeline-based supervised models (\citealp{nie2019combining}; \citealp{yoneda2018ucl}; \citealp{hanselowski2018ukp}).  To our knowledge, only \citealp{yoneda2018ucl} has provided the confusion matrix for each of the labels for their supervised system. For the same reason, we are only providing the comparison of the label accuracy on the ``SUPPORTS'' label in the results section.

\section{System Description}

In this section, we explain the design and all the underlying methods that our system has adopted. Our system is a pipeline consisting of three stages: (1) Question Generation, (2) Question Answering, (3) Label Classification. The question generation stage attempts to convert the claims into appropriate questions and answers. It generates questions similar to a Cloze-task or masked language modeling task where the named entities are masked with a blank. Question Answering stage predicts the masked blanks in an unsupervised manner. The respective predictions are then compared with the original answers and exported into a file for label classification. The label classifier calculates the predicted label based on a threshold.

\subsection{Question Generation}

The claims generally feature information about one or more entities. These entities can be of many types such as PERSON, CITY, DATE. Since the entities can be considered as the content words for the claim, we utilize these entities to generate the questions. Although function words such as conjunctions and prepositions form relationship between entities in the claims, we currently do not make use of such function words to avoid generating complex questions. The types of entities in a sentence can be recognized by using Stanford CoreNLP (\citealp{manning2014stanford}) NER tagger.

\begin{figure}[t]
\centering
\includegraphics[width=0.475 \textwidth]{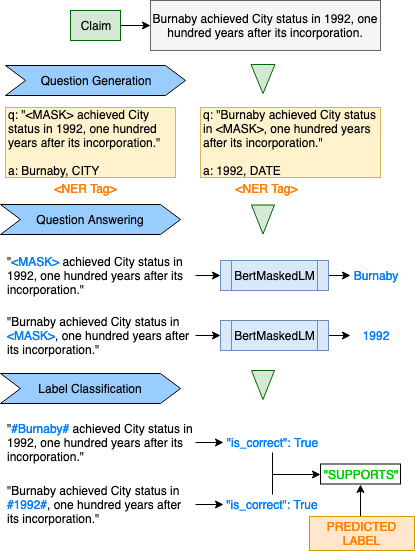}
\caption{An overview of the model pipeline}
\label{fig:process}
\end{figure}

\begin{table*}[!h]
\centering
\setlength{\tabcolsep}{4pt}
\begin{tabular}{| p{3cm} | p{2cm} | p{3cm} | p{2cm} | p{1.75cm} | p{2cm}|}
  \hline
  \textbf{Type of FEVER Set} & \textbf{Total Claims} & \textbf{Claims Converted to Questions} & \textbf{Conversion Accuracy} & \textbf{Total Questions} & \textbf{Questions per claim (Median)}\\
  \hline
  \hfil Training Set & \hfil145449 & \hfil131969 & \hfil90.73 & \hfil395717 & \hfil3\\
  \hline
  \hfil Development Set & \hfil19998 & \hfil17749 & \hfil88.75 & \hfil54422 & \hfil3\\
  \hline
  \hfil Test Set & \hfil9999 & \hfil8863 & \hfil88.63 & \hfil27359 & \hfil3\\
  \hline
\end{tabular}
\caption{Performance of the question generation system on FEVER Dataset.}
\label{tab1:performance}
\end{table*}

In our case, FEVER claims are derived from Wikipedia. We first collect all the claims from the FEVER dataset along with ``id'', ``label'' and ``verifiable'' fields. We don't perform any normalization on the claims such as lowercasing, transforming the spaces to underscore or parenthesis to special characters as it may decrease the accuracy of the NER tagger. These claims are then processed by the NER tagger to identify the named entities and their type. The named entities are then used to generate the questions by masking the entities for the subsequent stage. 

This process not only transforms the dataset but also transforms the task into a Cloze-task or masked language modeling task. Although the original masked language modeling task masks some of the tokens randomly, here we mask the named entities for generating the questions.

\subsection{Question Answering}
Originally inspired by the Cloze-task and developed to learn to predict the masked entities as well as the next sentence, BERT creates a deep bidirectional transformer model for the predictions. Since the FEVER claims are masked to generate the questions, we use BERT to tokenize the claims. We observed that the BERT tokenizer sometimes fails to tokenize the named entities correctly (e.g. Named entity \textit{“Taran”} was tokenized as \textit{``Tara'', ``\#\#n''}). This is due to the insufficient vocabulary used while training the WordPiece tokenizer.

To resolve this, we use Spacy Tokenizer\footnote{https://spacy.io/api/tokenizer} whenever the WordPiece Tokenizer fails. Once the claim is tokenized, we use the PyTorch Implementation of the BERT\footnote{https://github.com/huggingface/pytorch-transformers} model (BertForMaskedLM model) to predict the vocabulary index of the masked token. The predicted vocabulary index is then converted to the actual token. We compare the predicted token against the actual answer to calculate the label accuracy based on the classification threshold.

\subsection{Label Classification}

In this stage, we compute the final label based on the correctness score of the predictions that we received from the previous stage. The correctness score ($s$) is computed as:

\begin{equation}
\label{eq1:c_score}
s = \frac{n_c}{N}
\end{equation}

where $n_c$ indicates the number of correct questions, and $N$ is the total number of questions generated for the given claim. The label is assigned based on the correctness score ($s$) and the derived threshold ($\phi$) as:

\begin{equation}
\label{eq2:label}
    L(s)= 
    \begin{cases}
    \text{SUPPORTS},& \text{if } s\geq \phi\\
    \text{MANUAL\_REVIEW}, & \text{if } s < \phi
    \end{cases}
\end{equation}

Here, the classification threshold ($\phi$) is derived empirically based on the precision-recall curve.

\subsection{Model and Training details}
We utilize standard pre-trained BERT-Base-uncased model configurations as given below:
\begin{itemize}
    \item Layers: 12
    \item Hidden Units: 768
    \item Attention heads: 12
    \item Trainable parameters: 110M
\end{itemize}

We fine-tune our model (BERT) on the masked language modeling task on the wiki-text provided along with the FEVER dataset for 2 epochs.\footnote{In our experiments, after fine-tuning the model for 2 epochs there was no significant performance improvement.}

Note that Stanford CoreNLP NER tagger and the BERT model are the same for all the experiments and all the sets (development set, test set, training set).  We use the same PyTorch library mentioned in Section 2.2 for the fine-tuning as well.

\section{Results}

For the subtask of question generation, the results in Table \ref{tab1:performance} show that the system is able to generate questions given a claim with considerably good accuracy. The conversion accuracy is defined as the ratio of the number of claims in which the named entities are extracted to the number of claims. The results also support our assumption that the claims generally feature information about one or more entities.

Table \ref{tab2:accuracy} shows the performance of our Fact Checking system on the ``SUPPORTS'' label, the output of our system. We compare the results against two different classification thresholds. Table \ref{tab1:performance} shows that on an average there are 3 questions generated per claim. Here,  $\phi$ = 0.76 suggests that at least 3 out of the 4 questions have to be answered correctly while $\phi$ = 0.67 suggests that at least 2 out of the 3 questions has to be answered correctly for the claim to be classified as ``SUPPORTS''.
\begin{table}[h!]
\centering
\setlength{\tabcolsep}{4pt}
\begin{tabular}{| p{3cm} | p{1.9cm} | p{1.9cm} |}
  \hline
  \textbf{\hfil Type of Set} & \textbf{Label \newline Accuracy ($\phi$ = 0.76)} & \textbf{Label \newline Accuracy ($\phi$ = 0.67)} \\
  \hline
  Training Set & \hfil81.52 & \hfil88.05\\
  \hline
  Development Set & \hfil80.20 & \hfil86.7\\
  \hline
  Test Set & \hfil80.25 & \hfil87.04 \\
  \hline
\end{tabular}
\caption{Performance of the question generation system on FEVER Dataset.}
\label{tab2:accuracy}
\end{table}
  If only 1 question is generated, then it has to be answered correctly for the claim to be classified as ``SUPPORTS'' in case of both the thresholds.

In contrast to the results reported in Table \ref{tab2:accuracy}, here we consider $\phi$ = 0.76 to be a better classification threshold as it improvises over False Positives considerably over the entire dataset.

\begin{table}[h!]
\centering
\setlength{\tabcolsep}{4pt}
\begin{tabular}{| p{3cm} | p{1.9cm} | p{1.9cm} |}
  \hline
  \textbf{\hfil Model} & \textbf{ Label \newline Accuracy ($\phi$ = 0.76)} & \textbf{Label \newline Accuracy ($\phi$ = 0.67)} \\
  \hline
  HexaF - UCL & \hfil80.18 & \hfil80.18\\
  \hline
  Our Model (BERT) & \textbf{\hfil80.20} & \textbf{\hfil86.7}\\
  \hline
\end{tabular}
\caption{Comparison of the Label accuracy on Development set.}
\label{tab3:compare}
\end{table}

Although our unsupervised model doesn't support all the labels, to show the effectiveness of the approach, we compare the label accuracy of ``SUPPORTS'' label against a supervised approach -- HexaF. Results from Table \ref{tab3:compare} suggests that our approach is comparable to HexaF\footnote{Note that the label accuracy for HexaF is independent of the classification threshold $\phi$.} for $\phi$ = 0.76.

\section{Error Analysis}
\subsection{Question Generation}
The typical errors that we observed for the question generation system are due to the known limitations of the NER tagger. Most of the claims that the system failed to generate the questions from contain entity types for which the tagger is not trained. 

For instance, the claim ``A View to a Kill is an action movie.'' has a movie title (\textit{i.e. A View to a Kill}) and a movie genre (\textit{i.e. action}) but Stanford CoreNLP NER tagger is not trained to identify such type of entities.

\subsection{Question Answering}
We describe the most recurrent failure cases of our answering model in the description below.

\textbf{Limitations of Vocabulary.} Names like \textit{``Burnaby''} or \textit{``Nikolaj''} were not part of the original vocabulary while pre-training the BERT model, which makes it difficult to predict them using the same model. This was one of the most recurring error types.

\textbf{Limitations of Tokenizer.} The WordPiece tokenizer splits the token into multiple tokens. E.g. \textit{``Taran''} into \textit{``Tara'', ``\#\#n''}. In such cases, the answering system predicts the first token only which would be a substring of the correct answer. As we don't explicitly put a rule to avoid such cases, they are considered as incorrect answers. 

\section{Conclusion}
In this paper, we presented a transformer-based unsupervised question-answering pipeline to solve the fact checking task. The pipeline consisted of three stages: (1) Question Generation (similar to a Cloze-task), (2) Question Answering, (3) Label Classification. We use Stanford CoreNLP NER tagger to convert the claim into a Cloze-task by masking the named entities. The Question Generation task achieves almost 90\% accuracy in transforming the FEVER dataset into a Cloze-task. To answer the questions generated, we utilize masked language modeling approach from the BERT model. We could achieve 80.2\% label accuracy on ``SUPPORTS'' label. From the results, we conclude that it is possible to verify the facts with the right kind of factoid questions.

\section{Future Work}
To date, our approach only generates two labels ``SUPPORTS'' and ``MANUAL\_REVIEW''. We are working on extending this work to also generate ``REFUTED'' by improving our question generation framework. We will also work on generating questions using recent Neural Question Generation approaches. Later, to achieve better accuracy for tokenizing as well as answering, we plan to train the WordPiece Tokenizer from scratch.

\section*{Acknowledgments}

The authors thank Dr. Amit Nanavati and Dr. Ratnik Gandhi for their insightful comments, suggestions, and feedback. This research was supported by the TensorFlow Research Cloud (TFRC) program. \\

\bibliography{my_refs}
\bibliographystyle{bib_style}

\end{document}